\documentclass[runningheads]{llncs}

\usepackage{times}
\usepackage[misc]{ifsym}
\usepackage{epsfig}
\usepackage{graphicx}
\usepackage{amsmath, subfigure}
\usepackage{amssymb}
\usepackage{multicol,multirow}
\usepackage{cite}
\usepackage{booktabs}
\usepackage{bigstrut}
\usepackage{floatrow}
\usepackage{rotating}
\usepackage{adjustbox}
\usepackage{amsmath}

\newcommand{\mat}[1]{\boldsymbol{#1}}

\begin{document}

\title{Neural Network Compression by Joint Sparsity Promotion and Redundancy Reduction}

\titlerunning{Neural Network Compression by Joint Sparsity Promotion and Redundancy Reduction}

\author{Tariq M. Khan \Letter\inst{1}\and
Syed S. Naqvi\inst{2} \and
Antonio Robles-Kelly\inst{3} \and
Erik Meijering\inst{1}
}

\authorrunning{T. M. Khan et al.}

\institute{School of Computer Science and Engineering, UNSW, Australia \\
\email{tariq045@gmail.com}\and Department of Electrical and Computer Engineering, CUI, Pakistan \and Deakin University, Faculty of Science, Engineering and Built Environment,\\Waurn Ponds, VIC 3216, Australia}

\tocauthor
\toctitle

\maketitle

\begin{abstract}
Compression of convolutional neural network models has recently been dominated by pruning approaches. A class of previous works focuses solely on pruning the unimportant filters to achieve network compression. Another important direction is the design of sparsity-inducing constraints which has also been explored in isolation. This paper presents a novel training scheme based on composite constraints that prune redundant filters and minimize their effect on overall network learning via sparsity promotion. Also, as opposed to prior works that employ pseudo-norm-based sparsity-inducing constraints, we propose a sparse scheme based on gradient counting in our framework. Our tests on several pixel-wise segmentation benchmarks show that the number of neurons and the memory footprint of networks in the test phase are significantly reduced without affecting performance. MobileNetV3 and UNet, two well-known architectures, are used to test the proposed scheme. Our network compression method not only results in reduced parameters but also achieves improved performance compared to MobileNetv3, which is an already optimized architecture.
\end{abstract}

\section{Introduction}
One of the key enablers of the unprecedented success of deep learning in recent years is the availability of very large neural network models \cite{khan2020machine}. Modern deep neural networks (DNNs), in particular convolutional neural networks (CNNs), typically consist of many cascaded layers totaling millions to hundreds of millions of parameters (weights) \cite{NIPS2012_c399862d, pmlr-v48-linb16,khan2021leveraging}. The larger-scale neural networks tend to enable the extraction of more complex high-level features and therefore lead to a significant improvement of the overall accuracy \cite{6639343,6248110,SCHMIDHUBER201585}. On the other side, the layered deep structure and large model sizes also demand increasing computational and memory resources\cite{khan2022t,khan2022width}. The problem with many modern neural networks is that they suffer from parameter explosion \cite{khan2021residual,khan2021rc}. Due to the large number of parameters, training can take weeks on a central processing unit (CPU) or days on a graphics processing unit (GPU) \cite{khan2020derivative}. For larger and higher-dimensional image sizes, even a GPU is not able to train a network with a large number of parameters.

The aim of the work presented in this paper is to design sparse, low-complexity neural networks by reducing the number of parameters while keeping performance degradation negligible. Memory and computational requirements in particular complicate the deployment of deep neural networks on low-power embedded platforms as they have limited computing and power budget. The energy efficiency challenge of large models motivates model compression. Several algorithm-level techniques have been proposed to compress models and accelerate DNNs, such as quantization to 16-bit \cite{gupta15}, group L1 or L2 regularization \cite{8600728}, node pruning \cite{572092,5564272,Wen2016}, filter pruning for CNNs (\cite{li2017pruning,Wen2016,8237417,8237560}, weight pruning using magnitude-based methods \cite{NIPS2015_ae0eb3ee}, weight quantization \cite{7780890, pmlr-v48-linb16}, connection pruning \cite{NIPS2015_ae0eb3ee}, and low-rank approximation \cite{tai2016convolutional}.

Pruning--–which removes entire filters, or neurons, that make little or no contribution to the output of a trained network---is a way to make a network smaller and faster. There are two forms in which structural pruning is commonly applied: i) using a predefined per-layer pruning ratio, or ii) simultaneously over all layers. The second form allows pruning to automatically find a better architecture [24]. An exact solution for pruning will be to minimize the $l_0$ norm of all neurons and remove those that are zeroed out. However, $l_0$ minimization is impractical as it is non-convex, NP-hard, and requires combinatorial search. Therefore, prior work has tried to relax the optimization using Bayesian methods [25, 29] or regularization terms.

Motivated by the success of sparse coding, several methods relax $l_0$ minimization with $l_1$ or $l_2$ regularization, followed by soft thresholding of parameters with a predefined threshold. These methods belong to the family of iterative shrinkage and thresholding algorithms (ISTA) [3]. Han et al. [7] applied a similar approach for removing individual weights of a neural network to obtain sparse non-regular convolutional kernels. Li et al. [22] extended this approach to remove filters with small $l_1$ norms.

Due to the popularity of batch-normalization [17] layers in recent networks [10, 15], several approaches have been
proposed for filter pruning based on batch-norm parameters [23, 33]. These works regularize the scaling term ($\gamma$) of batch-norm layers and apply soft thresholding when the value falls below a predefined threshold. Furthermore, floating-point operations (FLOPS) based penalties can also be included to directly reduce computational costs [5]. A more general scheme that uses an ISTA-like method on scaling factors [16] can be applied to any layer.

While these approaches can offer a reasonable parameter reduction (e.g.\ by 9$\times$ to 13$\times$ in \cite{NIPS2015_ae0eb3ee}) with minor accuracy degradation, they suffer from three drawbacks: 1) the sparsity regularization and pruning typically result in an irregular network structure, thereby undermining the compression ratio and limiting performance and throughput \cite{8192500}; 2) the training complexity is increased due to the additional pruning process \cite{NIPS2015_ae0eb3ee} or low-rank approximation step \cite{tai2016convolutional}; and 3) the compression ratios depending on network are heuristic and cannot be precisely controlled.

The main contributions of this work are threefold. First, we propose a novel training scheme based on composite constraints that removes redundant filters during training. Second, our experiments on several benchmarks demonstrate a significant reduction in the number of neurons and the memory footprint of networks in the test phase without affecting their accuracy. And, finally, we demonstrate online filter redundancy removal in encoder-decoder networks for the task of pixel-wise segmentation.

\section{Methodology}
\subsection{Loss Function}
Our goal is to formulate the loss function with added constraints given as
\begin{equation}\label{eqn:loss}
f = \alpha\left({y,y'}\right) + g(\mat{W}),
\end{equation}
where $\alpha$ is the standard loss, $y$ and $y'$ are the true and predicted labels respectively, and $g(\mat{W})$ represents the set of added constraints on the network parameters $\mat{W}$. Correspondingly, the $l$-th iteration of the back-propagation can be defined as:
\begin{equation}\label{eqn:weightUpdate}
\mat{W}_{l}^{t + 1} = \mat{W}_{l}^{t} + \gamma \nabla_{{\mat{W}_{l}}}\alpha + \gamma\nabla\left({\sum_{i \in l}}{g\left({{\mat{W}_i}}\right)}\right)\!,
\end{equation}
where $\mat{W}_{l}$ represents the parameters learned in the $l$-th iteration, $t$ and $t+1$ represent the current and next states, and $\gamma$ is the learning rate. According to (\ref{eqn:weightUpdate}), a new term $\nabla\left( {\sum_{i \in l}} {g\left( {{\mat{W}_i}} \right)}  \right)$ must be added to each constrained layer during back-propagation to compute the overall gradient. In our case, the two terms that define $g(\mat{W})$ are differentiable and therefore can be easily incorporated in the back-propagation as shown in later sections.

\subsection{Composite Constraints}
We aim to remove the effect of redundant filters during training and effectively minimize network parameters without compromising network performance. To this end, we propose the following two composite constraints for $g(\mat{W})$ in (\ref{eqn:loss}).\\

\textbf{Inter-Filter Orthogonality Constraint:} The idea of inter-filter orthogonality has been explored in the past \cite{Prakash2019} as a measure to guide the filter pruning process, which is generally performed in an offline manner. In contrast, we employ it as a regularization term in our loss function to remove the effect of redundant filters during online training. The proposed inter-filter orthogonality constraint $R$ is given as:
\begin{equation}\label{eqn:IFOC}
R\left( \mat{W} \right) = \left\| {\mat{W}{\mat{W}^T} - I} \right\|_2^2.
\end{equation}
Here, instead of using the absolute difference \cite{Prakash2019}, we use the squared $l_2$ norm, which we found to perform well in our experiments.\\

\textbf{Sparse Regularization Constraint:} Prior works have explored group-sparse regularization including $l_2$ or $l_1$ norms \cite{Zhou2016,Wen2016}. Instead, we propose to use the $l_0$ norm to further eliminate the contribution of redundant filters towards network learning. The proposed sparsity constraint is defined as:
\begin{equation}\label{eqn:SRC}
H\left(\mat{S}_{l}\right) = {\left\|\mat{S}_{l}\right\|_0}.
\end{equation}
Here, $\mat{S}_l$ represent the layer-wise sparse gradients corresponding to the weights.

\subsection{Overall Update Equations}
Having defined the composite constraints constituting $g(\mat{W})$, we now define the overall update equations during online training. The updated loss function incorporating the composite constraints is defined as:
\begin{equation}\label{eqn:finalLoss}
f = \alpha \left( {y,y'} \right) + \lambda \sum\limits_{i \in l} {R\left( {{\mat{W}_i}} \right) + \beta } \sum\limits_{i \in l} {H\left( {{\mat{S}_i}} \right)},
\end{equation}
where $\lambda$ and $\beta$ are the parameters controlling the two regularization terms. Correspondingly, the weight update in the $l$-th iteration incorporating the gradient terms for the constraints is given as:
\begin{equation}\label{eqn:updatedweightUpdate}
\mat{W}_l^{t + 1} = \mat{W}_l^t + \gamma \,{\nabla _{{\mat{W}_{l}^{t}}}}\alpha  + \lambda \gamma \,\nabla \left( {\sum\limits_{i \in l} {R\left( {{\mat{W}_i}} \right)} } \right) + \beta \gamma \,\nabla \left( {\sum\limits_{i \in l} {H\left( \mat{S}_i \right)} } \right)\!.
\end{equation}
As the contribution of the sparsity term approaches zero,
\begin{equation}
\beta \gamma \,\nabla \left( {\sum\limits_{i \in l} {H\left( \mat{S}_i \right)} } \right) = 0,
\end{equation}
the update equation (\ref{eqn:updatedweightUpdate}) becomes
\begin{equation}\label{eqn:finalweightUpdate}
\mat{W}_l^{t + 1} = \mat{W}_{l}^{t} + \gamma \,{\nabla _{\mat{W}_{l}^{t}}}\alpha  + \tau \,{\nabla _{{\mat{W}_{l}^{t}}}}R\left( {{\mat{W}_l}} \right),
\end{equation}
where the gradient for the inter-filter orthogonality constraint is calculated as:
\begin{equation}
\nabla _{{\mat{W}_l}}R\left( {{\mat{W}_l}} \right) = \mat{W}{\left\| {\mat{W}{\mat{W}^T} - I} \right\|_2}.
\end{equation}
The update for the sparsity regularization term is obtained as:
\begin{equation}
\mat{S}_l^{t + 1} = \mat{S}_l^t + \gamma {\nabla _{\mat{S}_l^t}}\alpha  + \zeta \,{\nabla _{\mat{S}_l^t}}H\left( {\mat{S}_l^t} \right).
\end{equation}









\section{Experiments}

\subsection{Benchmark Datasets}
We have evaluated our proposed training strategy on three publicly available benchmark datasets across diverse applications including medical image segmentation, saliency detection, and scene understanding. Specifically, we employed the retinal vessel segmentation dataset DRIVE~\cite{DRIVE}, the salient object dataset DUT-OMRON~\cite{DUT-OMRON}, and the video recognition dataset CamVid~\cite{CamVid} for these applications, respectively. The DRIVE~\cite{DRIVE} dataset consists of 40 retinal scans from a population of 400 diabetic patients from the Netherlands, out of which only 7 exhibited signs of mild diabetic retinopathy. The color images (RGB) were acquired using a Canon 3CCD camera at a resolution of $768\times584$ pixels at 8 bits per pixel per color channel and are stored in JPEG format. Half of the images constitute the training set, while the other half were employed as test images. Manual segmentations of the vasculature verified by experts are available as gold standard for both training and test images. The standard training and test sets were employed in our experiments. The CamVid~\cite{CamVid} dataset includes 700 color images (RGB) with pixel-wise reference masks for 32 semantic classes. As the name suggests, the images were derived from 30 Hz footage of driving scenarios acquired from CCTV cameras from the perspective of a driving automobile. In this work, 307 training images, 60 validation images, and 101 test images with labels corresponding to three semantic classes were employed. The included classes feature the ``sky'' class and ``building'' class while all other semantic objects are grouped into a single ``other'' class. The images have a resolution of 360$\times$480 pixels and are stored in PNG format. Finally, the DUT-OMRON~\cite{DUT-OMRON} dataset consists of a collection of 5168 natural color images (RGB) with various resolutions available in JPEG format. The images include one or more salient objects with a challenging background. Binary reference masks are also available. For our experiments, we randomly sampled 420 images from the DUT dataset, with 70-30 ratio for training and test images.

\subsection{Network Models}
To demonstrate the prunability of the proposed scheme, we selected U-Net~\cite{U-Net} as a large-scale model and MobileNetV3-Small\cite{mobileNet} as a lightweight model for our experiments. For U-Net, we employed the implementation from \cite{Kezmann:2020}\footnote{\url{https://github.com/JanMarcelKezmann/TensorFlow-Advanced-Segmentation-Models}}, whereas the sub-classing-based implementation of \cite{Larry:2020}\footnote{\url{https://github.com/xiaochus/MobileNetV3}} was adopted for MobileNetV3-Small. We used weight-based pruning in our experiments, where 10\% and 50\% weights are magnitude pruned. To evaluate the efficacy of the proposed scheme on a wide range of weight percentages, we used iterative pruning. Specifically, we used three iterations of 10\% and 50\% pruning, and after each iteration the model was trained for an additional 5 epochs to update the weights. The experimental results include performance results for different iterations specified by the keyword ``iter''. For training the networks, stochastic gradient descent with a learning rate of 0.2 and a momentum of 0.9 was used, where the learning rate was reduced on plateaus. The models were first trained for 100 epochs before pruning with early stopping. The intersection-over-union (IOU) score was employed for monitoring the learning rate and early stopping.

\subsection{Evaluation Measures}\label{EM}
The available gold-standard reference segmentation maps are binary images indicating for each pixel whether it corresponds to a feature (object) or non-feature (background). Hence there are four types of pixels for each output image: feature pixels correctly predicted as features (true positives: TP), non-feature pixels correctly predicted as non-features (true negatives: TN), non-feature pixels incorrectly predicted as features (false positives: FP), and feature pixels incorrectly predicted as non-features (false negatives: FN). From these, we calculated several common performance metrics \cite{metricsreloaded}:

\begin{equation}
\label{Sens}
\text{Sensitivity} = \frac{\text{TP}}{\text{TP} + \text{FN}},
\end{equation}

\begin{equation}
\label{Spec}
\text{Specificity} = \frac{\text{TN}}{\text{TN} + \text{FP}},
\end{equation}

\begin{equation}
\label{Acc}
\text{Accuracy} = \frac{\text{TP} + \text{TN}}{\text{TP} + \text{FN} + \text{TN} + \text{FP}},
\end{equation}

\begin{equation}
\label{BAcc}
\text{Balance Accuracy} = \frac{\text{Sensitivity} + \text{Specificity}}{2},
\end{equation}

\begin{equation}
\label{F1}
\text{F1} = \frac{\text{2TP}}{\text{2TP} + \text{FP} + \text{FN}},
\end{equation}

\begin{equation}
\label{Jaccard}
\text{Jaccard} = \frac{\text{TP}}{\text{TP} + \text{FP} + \text{FN}},
\end{equation}

\begin{equation}
\label{Error}
\text{Error} = 1 - \text{Jaccard}.
\end{equation}
The Jaccard measure \eqref{Jaccard} is the ratio between the intersection and the union of the segmented image and the gold-standard mask and is also known as the intersection-over-union (IOU) measure \cite{Everingham2014ThePV}. For all measures, higher values imply better performance, except for the error measure \eqref{Error} where lower values imply better performance.

\begin{table*}[!b]
  \centering
  \caption{Performance comparison of the standard loss and the proposed custom loss for the MobileNetV3-Small network on the DRIVE dataset. Bold indicates best performance per metric.}
  	\resizebox{1\textwidth}{!}{
    \begin{tabular}{@{}l@{\hspace{1em}}c@{\hspace{1em}}c@{\hspace{1em}}c@{\hspace{1em}}c@{\hspace{1em}}c@{\hspace{1em}}c@{\hspace{1em}}c@{\hspace{1em}}c@{}}
    \toprule
    \textbf{Method} & \textbf{Sens.} & \textbf{Spec.} & \textbf{Acc.} & \textbf{BAcc.} & \textbf{F1} & \textbf{Jacc.} & \textbf{Error} \\
    \midrule
    MobileNetV3-Scratch                       & \bf 0.6054 & 0.9486 & 0.9186 & \bf 0.7770 & 0.5652 & 0.3945 & 0.6055 \\
    MobileNetV3-CustomLoss-Scratch            & 0.5860 & \bf 0.9554 & \bf 0.9231 & 0.7707 & \bf 0.5709 & \bf 0.4001 & \bf 0.5999 \\
    \midrule
    MobileNetV3-Scratch-AddFilters            & 0.6264 & 0.9521 & 0.9236 & 0.7892 & 0.5885 & 0.4175 & 0.5825 \\
    MobileNetV3-CustomLoss-Scratch-AddFilters & \bf 0.6381 & \bf 0.9585 & \bf 0.9305 & \bf 0.7983 & \bf 0.6158 & \bf 0.4453 & \bf 0.5547 \\
    \midrule
    MobileNetV3-Pretrained                    & \bf 0.6480 & 0.9593 & 0.9322 & 0.8037 & 0.6250 & 0.4550 & 0.5450 \\
    MobileNetV3-CustomLoss-Pretrained         & 0.6477 & \bf 0.9614 & \bf 0.9341 & \bf 0.8046 & \bf 0.6315 & \bf 0.4619 & \bf 0.5381 \\
    \bottomrule
    \end{tabular}
    }
  \label{tab:losscomparison}
\end{table*}

\begin{table*}[!t]
  \centering
  \caption{Parameter saving ability of the proposed loss. Test times were computed without removing the saved parameters.}
    \begin{tabular}{@{}l@{\hspace{3em}}c@{\hspace{1em}}c@{\hspace{1em}}c@{\hspace{1em}}c@{}}
    \toprule
    \textbf{Method} & \textbf{Parameters} & \textbf{Parameters} & \textbf{Train} & \textbf{Test} \\
    & \textbf{Backbone} & \textbf{Saved} & \textbf{Time} & \textbf{Time} \\
    \midrule
    MobileNetV3-Pretrained                   & 0.445M &        & 777 & 0.107 \\
    MobileNetV3-Pretrained-Optimized         & 0.433M & 11,290 & 766 & 0.105 \\
    \midrule
    MobileNetV3-Scratch-AddFilters           & 0.531M &        & 760 & 0.109 \\
    MobileNetV3-Scratch-AddFilters-Optimized & 0.517M & 13,035 & 753 & 0.104 \\
    \bottomrule
    \end{tabular}
  \label{tab:parameterSaving}
\end{table*}

\begin{figure*}[!b]
	\centering
		\begin{tabular}{@{}cccccccc@{}}
			C1 & C2 & C3 & C4 & C5 & C6 & C7 & C8 \\
			\subfigure{\includegraphics[scale=0.072]{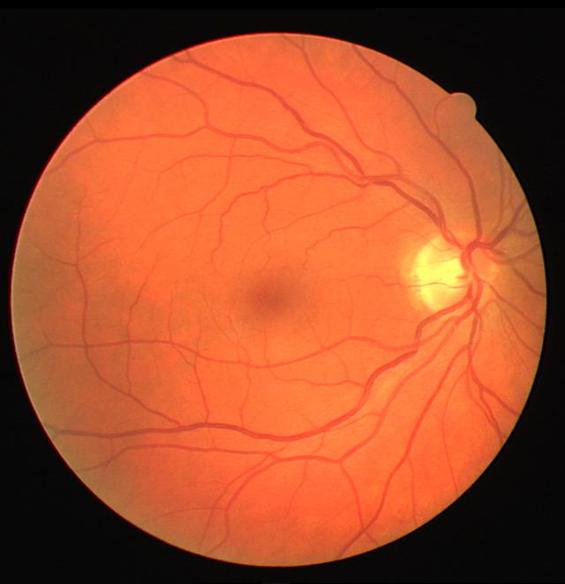}} &
			\subfigure{\includegraphics[scale=0.072]{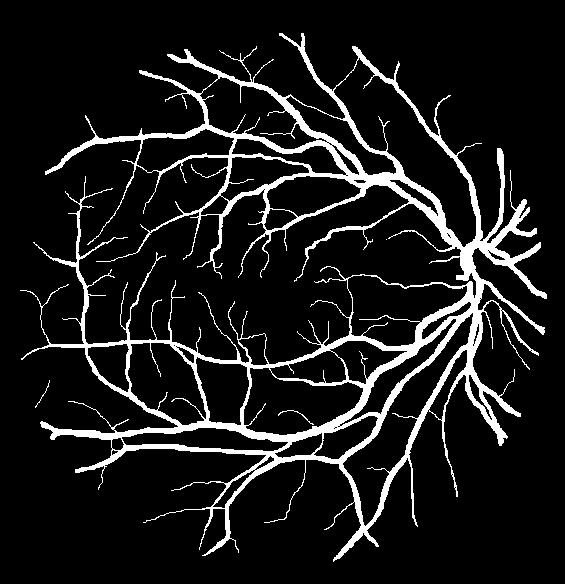}} &
		    \subfigure{\includegraphics[scale=0.072]{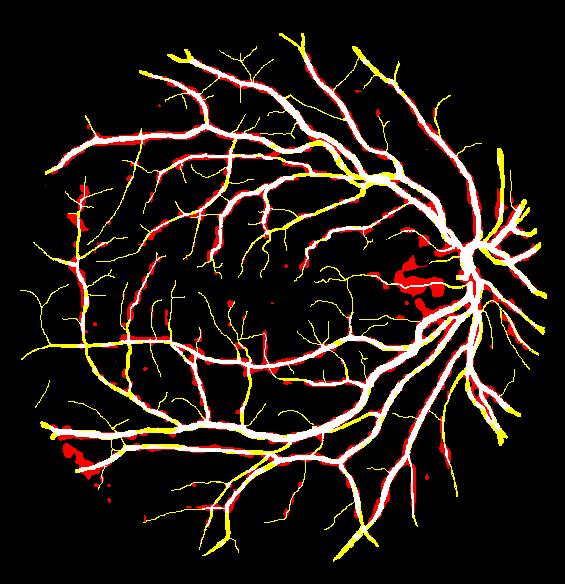}} &
		    \subfigure{\includegraphics[scale=0.072]{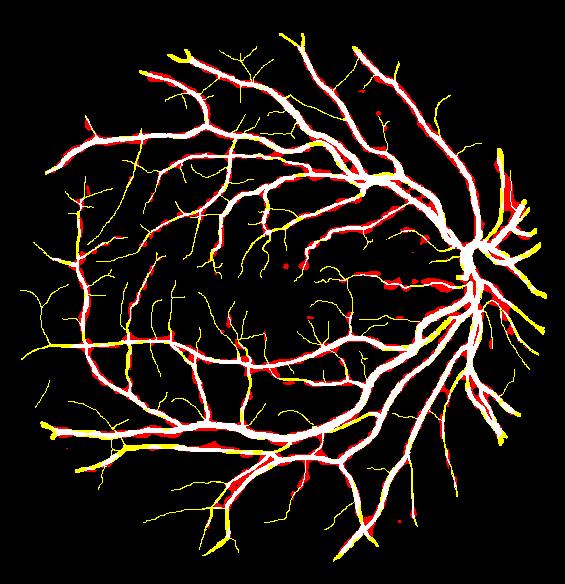}} &
		    \subfigure{\includegraphics[scale=0.072]{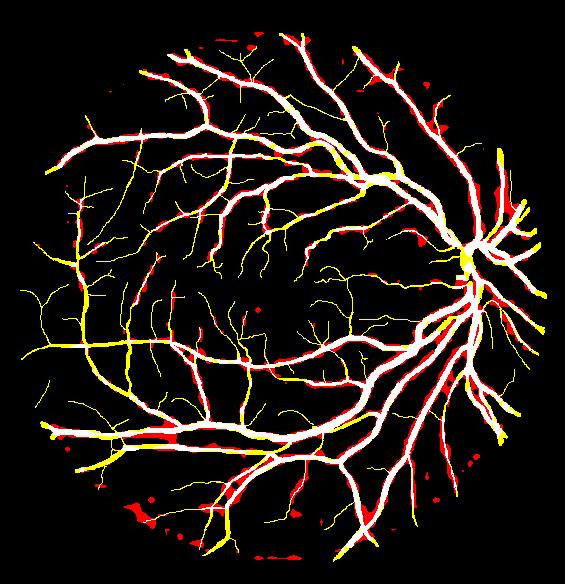}} &
		    \subfigure{\includegraphics[scale=0.072]{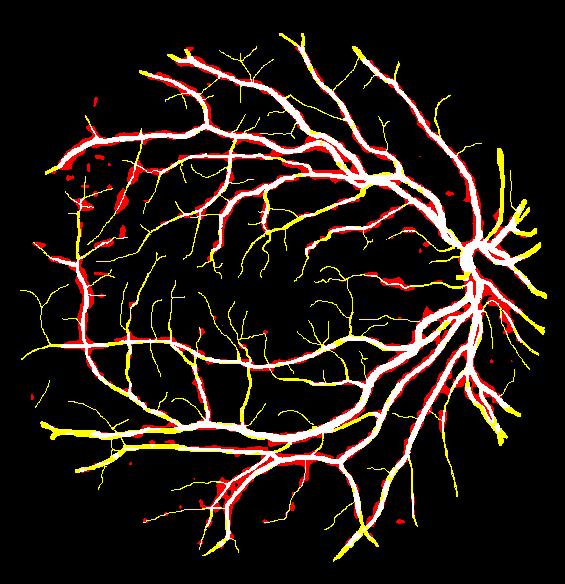}} &
		    \subfigure{\includegraphics[scale=0.072]{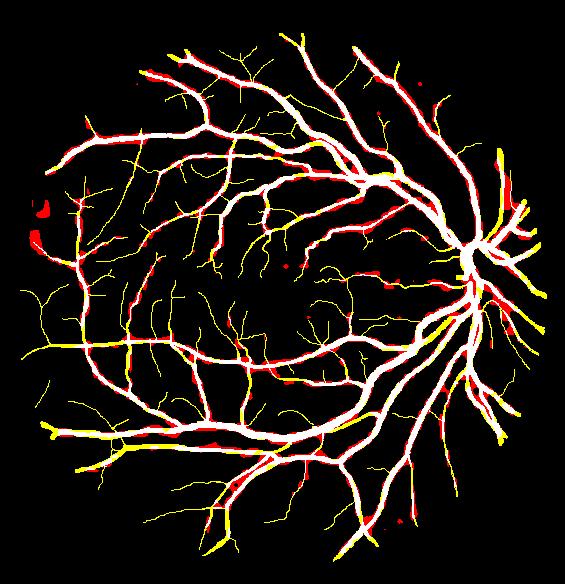}} &
	        \subfigure{\includegraphics[scale=0.072]{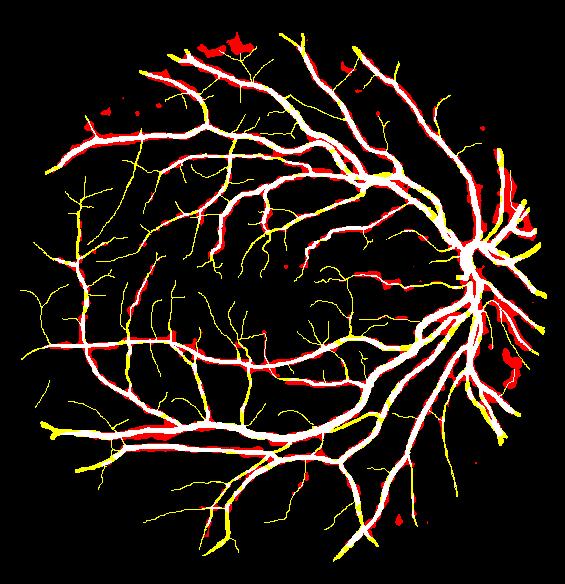}}\\[-1mm]
	        \subfigure{\includegraphics[scale=0.072]{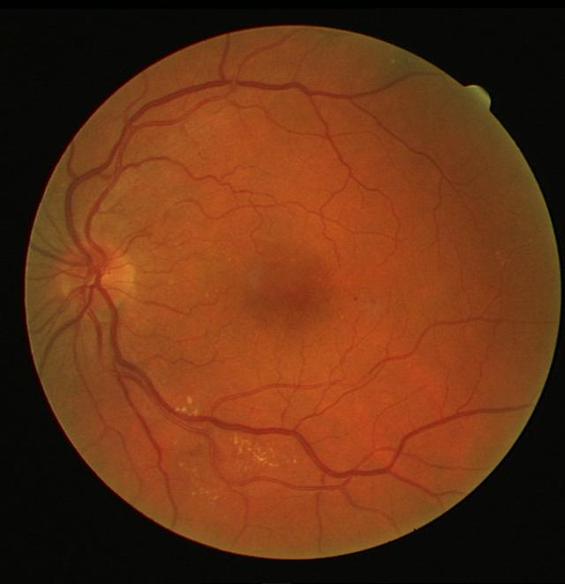}} &
			\subfigure{\includegraphics[scale=0.072]{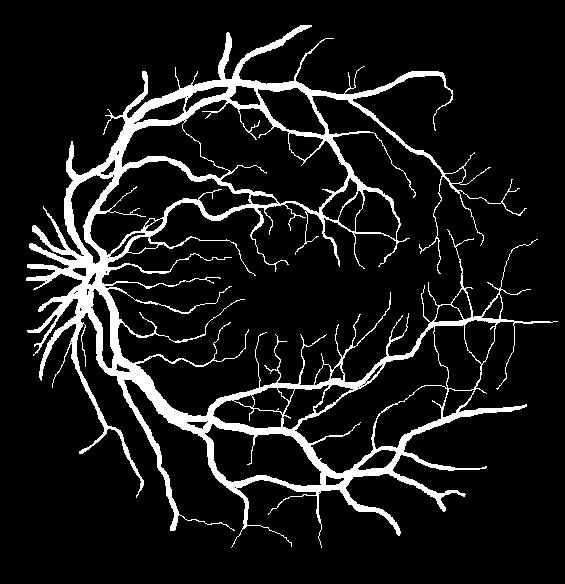}} &
		    \subfigure{\includegraphics[scale=0.072]{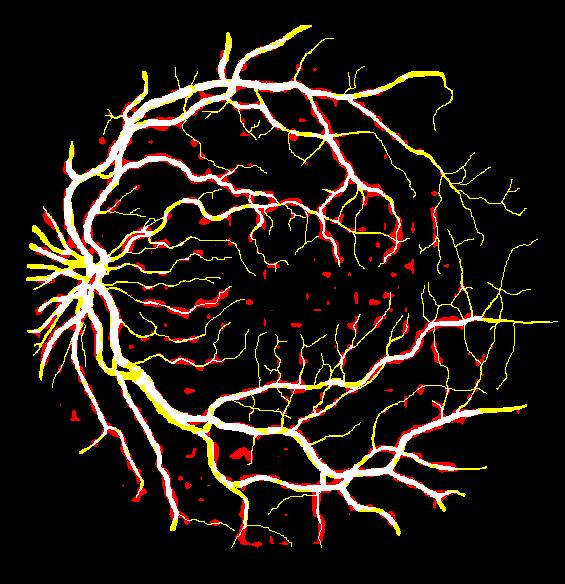}} &
		    \subfigure{\includegraphics[scale=0.072]{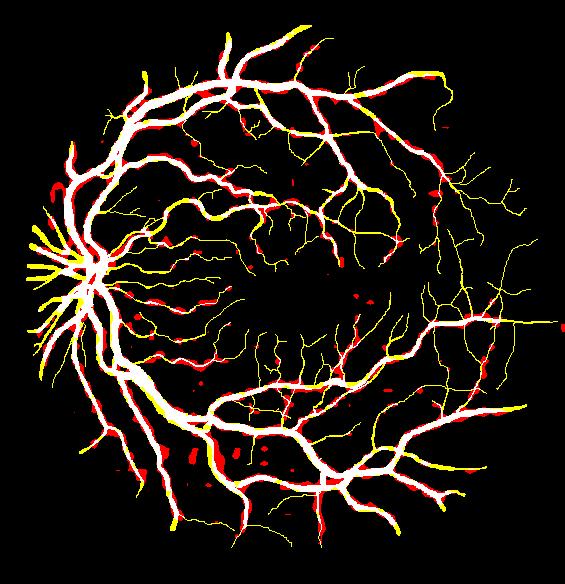}} &
		    \subfigure{\includegraphics[scale=0.072]{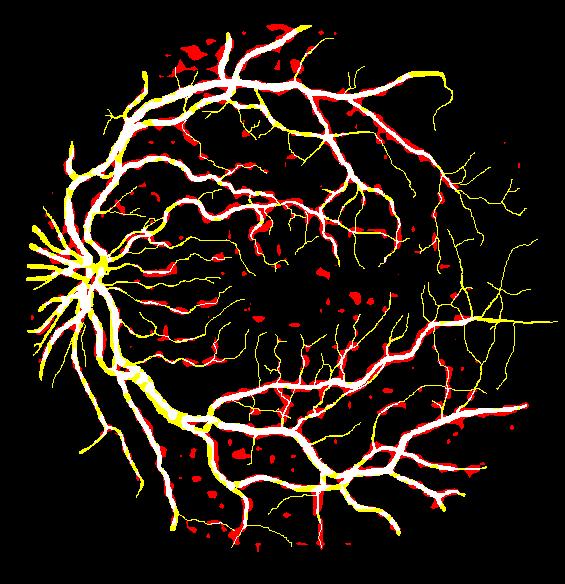}} &
		    \subfigure{\includegraphics[scale=0.072]{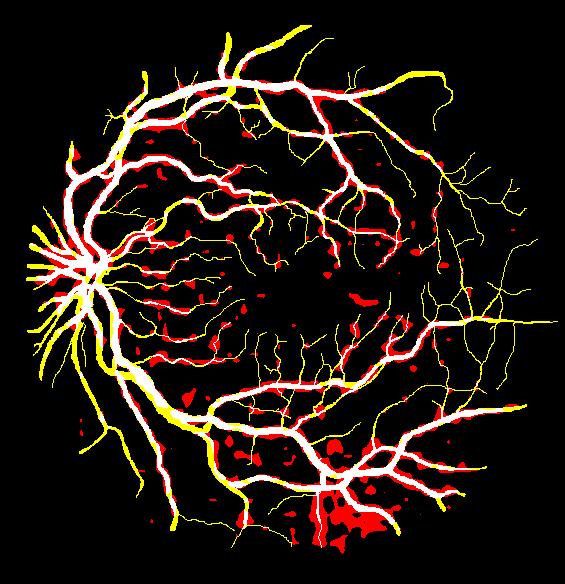}} &
		    \subfigure{\includegraphics[scale=0.072]{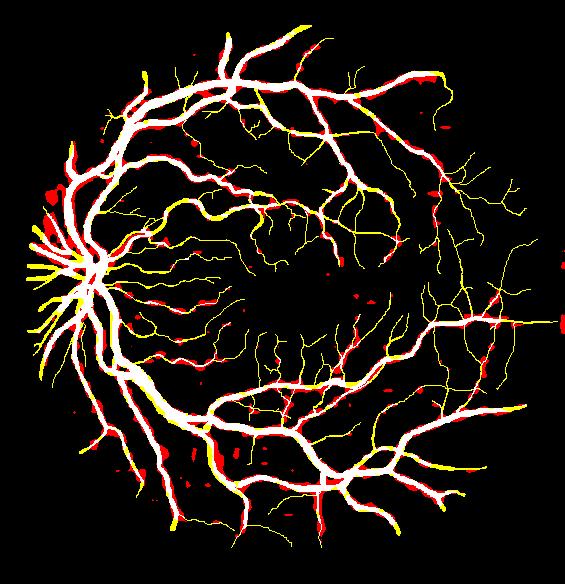}} &
	        \subfigure{\includegraphics[scale=0.072]{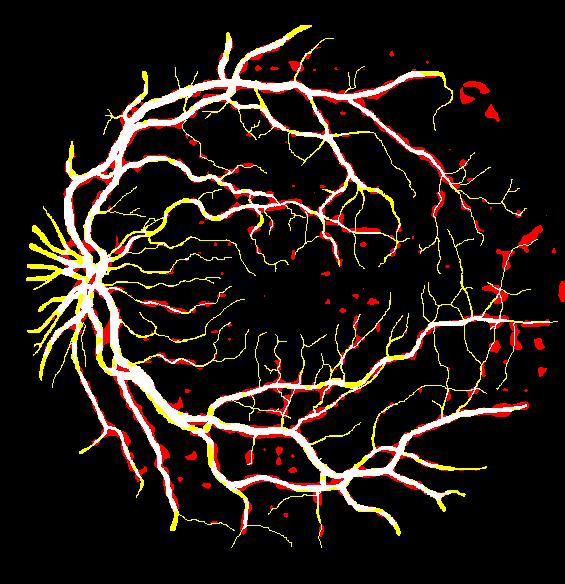}}\\[-1mm]
	        \subfigure{\includegraphics[scale=0.072]{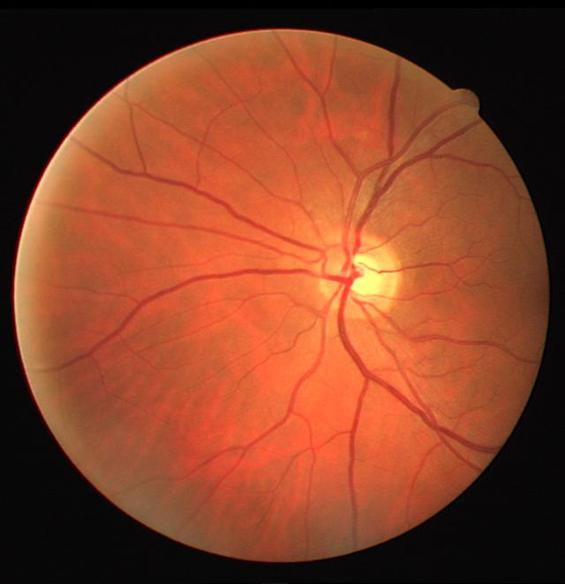}} &
			\subfigure{\includegraphics[scale=0.072]{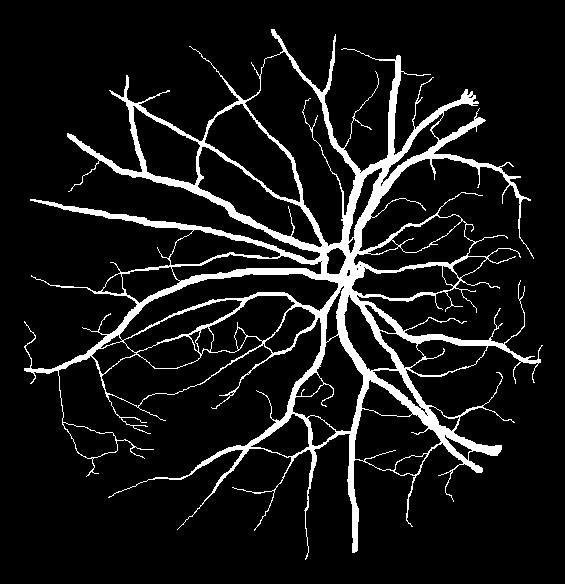}} &
		    \subfigure{\includegraphics[scale=0.072]{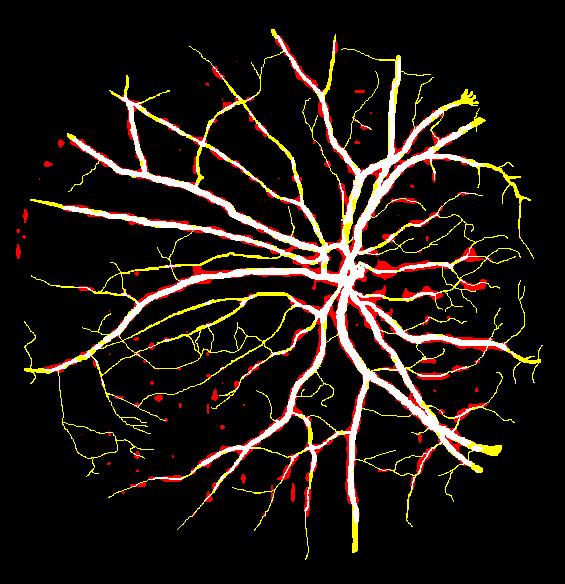}} &
		    \subfigure{\includegraphics[scale=0.072]{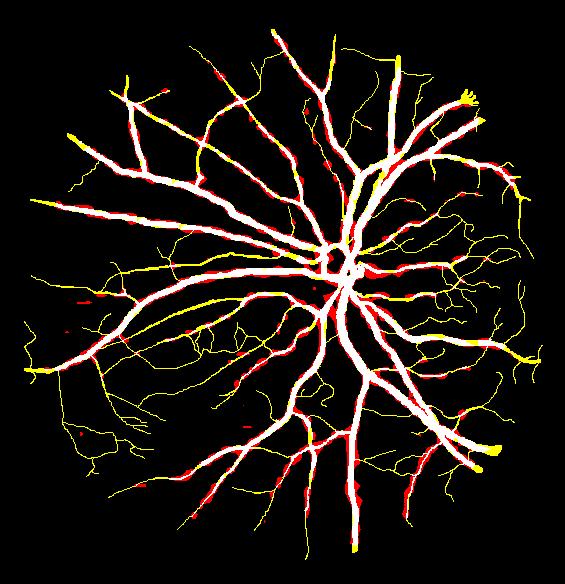}} &
		    \subfigure{\includegraphics[scale=0.072]{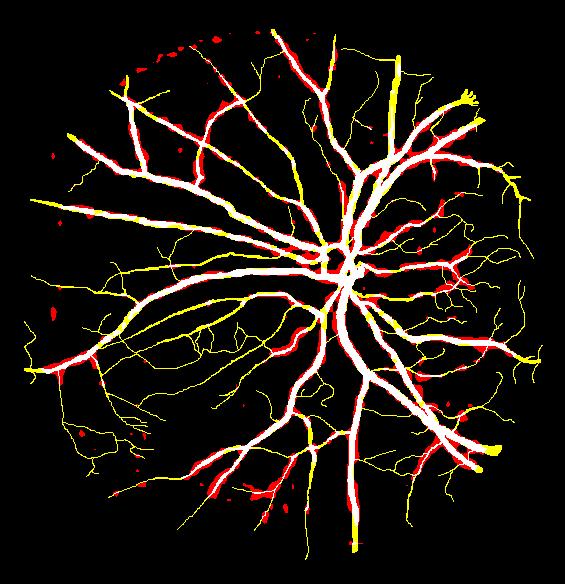}} &
		    \subfigure{\includegraphics[scale=0.072]{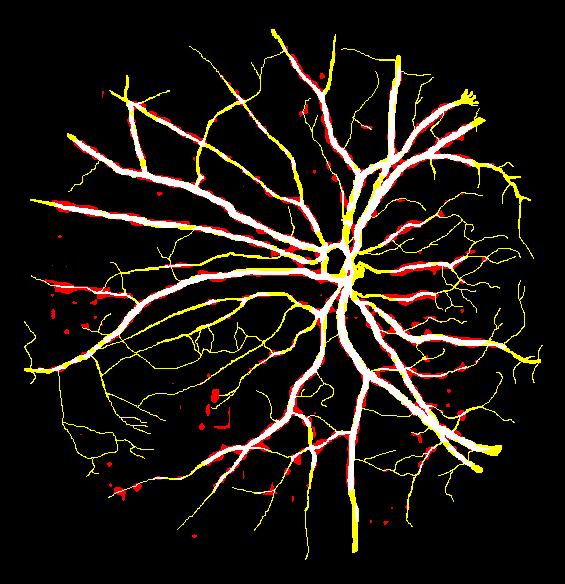}} &
		    \subfigure{\includegraphics[scale=0.072]{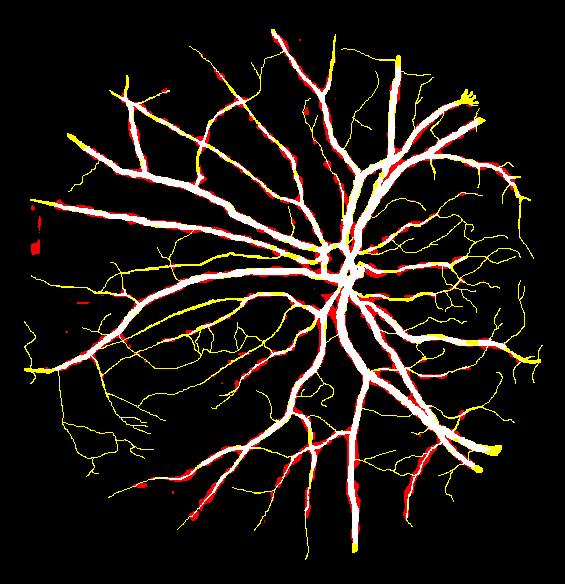}} &
	        \subfigure{\includegraphics[scale=0.072]{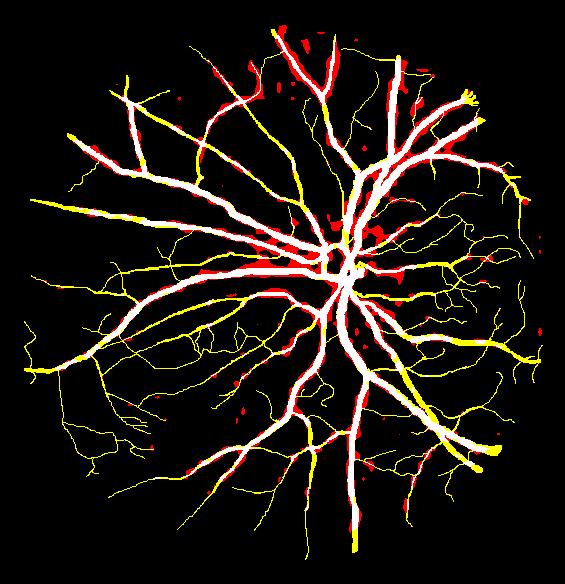}}\\[-1mm]
	        \subfigure{\includegraphics[scale=0.072]{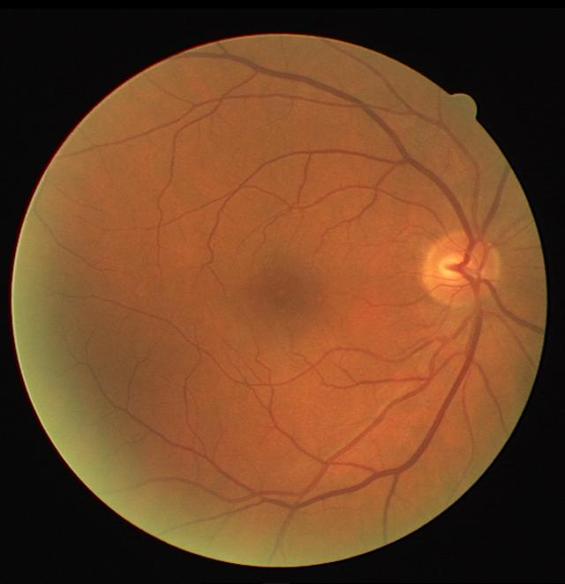}} &
			\subfigure{\includegraphics[scale=0.072]{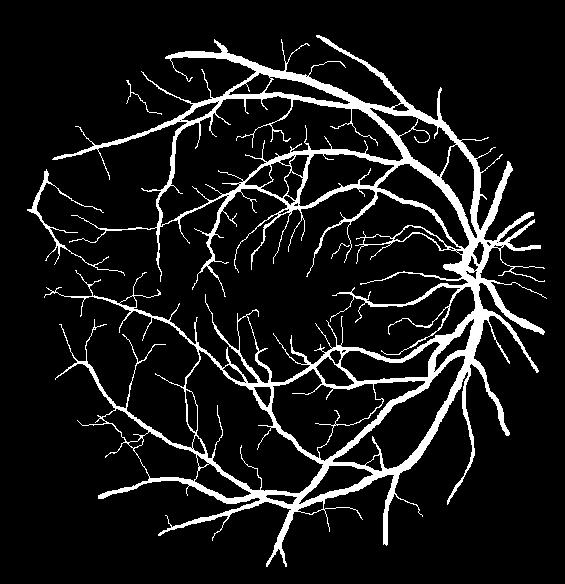}} &
		    \subfigure{\includegraphics[scale=0.072]{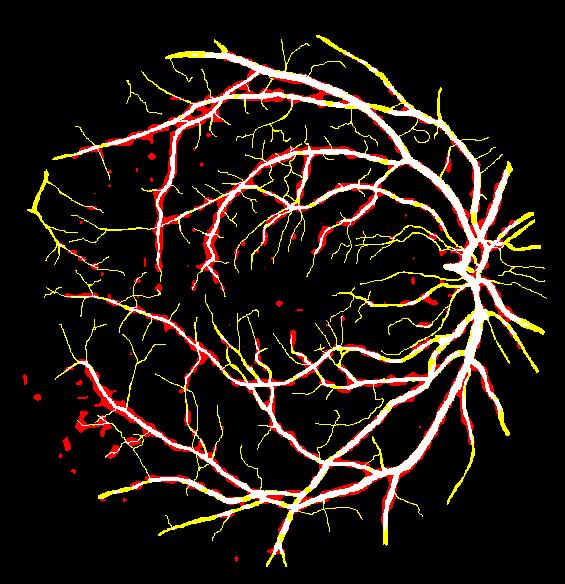}} &
		    \subfigure{\includegraphics[scale=0.072]{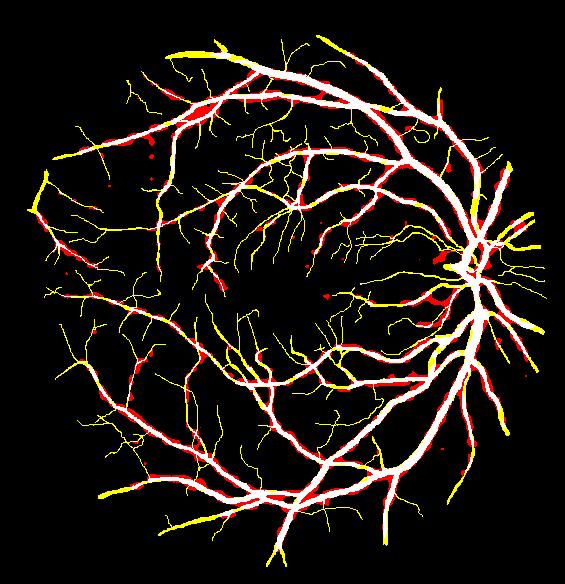}} &
		    \subfigure{\includegraphics[scale=0.072]{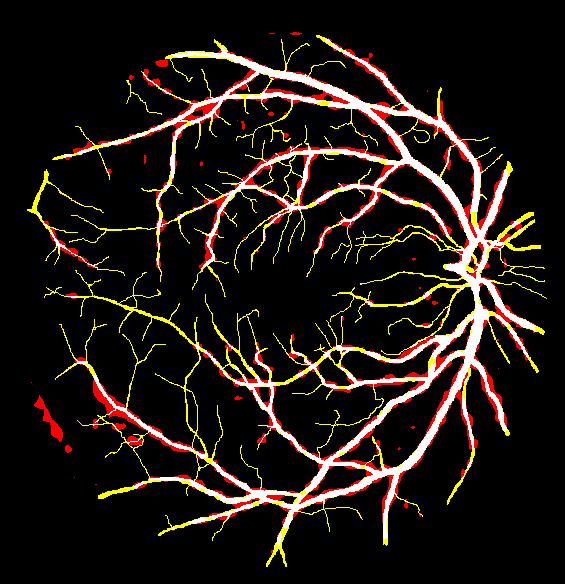}} &
		    \subfigure{\includegraphics[scale=0.072]{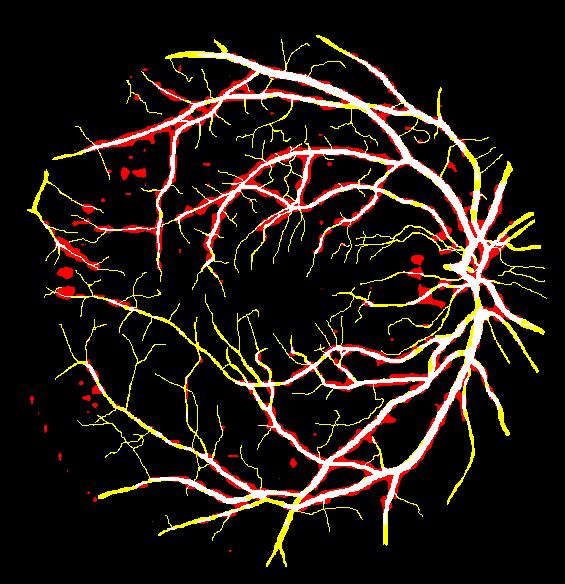}} &
		    \subfigure{\includegraphics[scale=0.072]{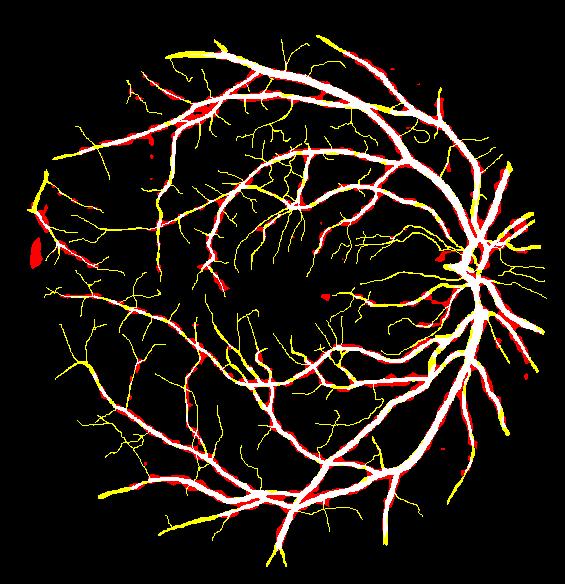}} &
	        \subfigure{\includegraphics[scale=0.072]{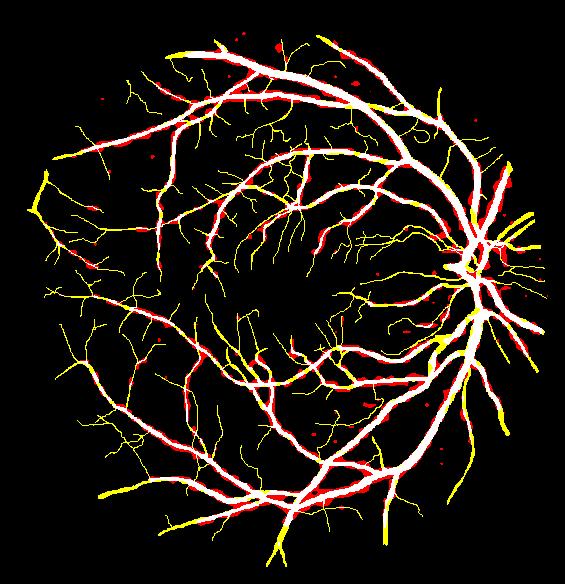}}\\[-1mm]
	        \subfigure{\includegraphics[scale=0.072]{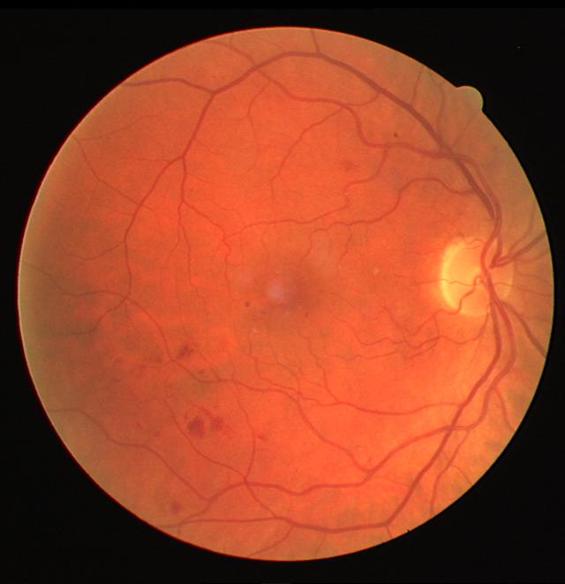}} &
			\subfigure{\includegraphics[scale=0.072]{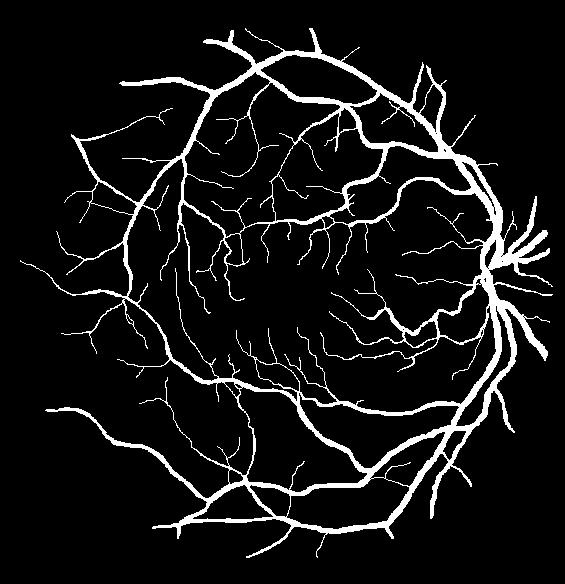}} &
		    \subfigure{\includegraphics[scale=0.072]{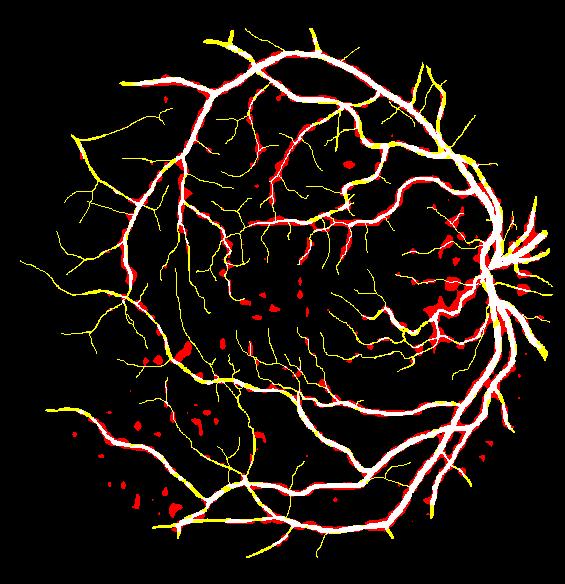}} &
		    \subfigure{\includegraphics[scale=0.072]{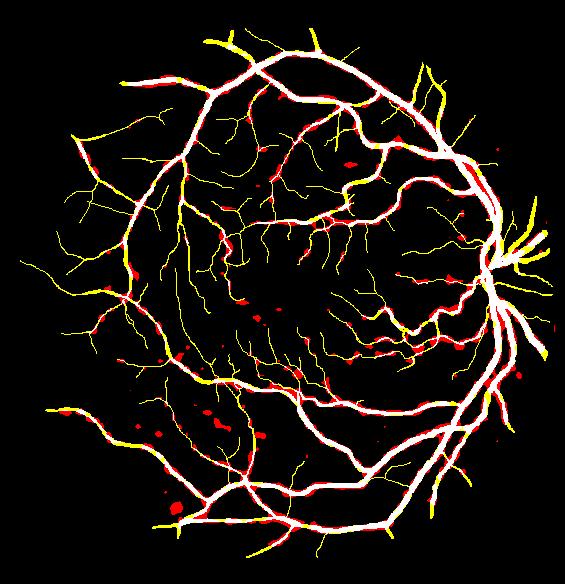}} &
		    \subfigure{\includegraphics[scale=0.072]{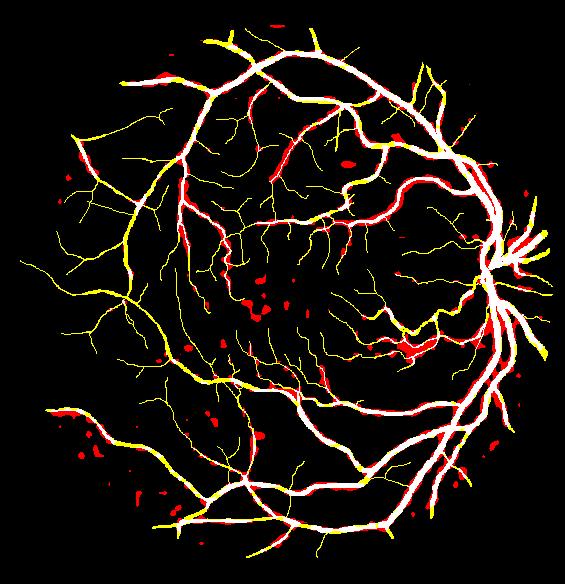}} &
		    \subfigure{\includegraphics[scale=0.072]{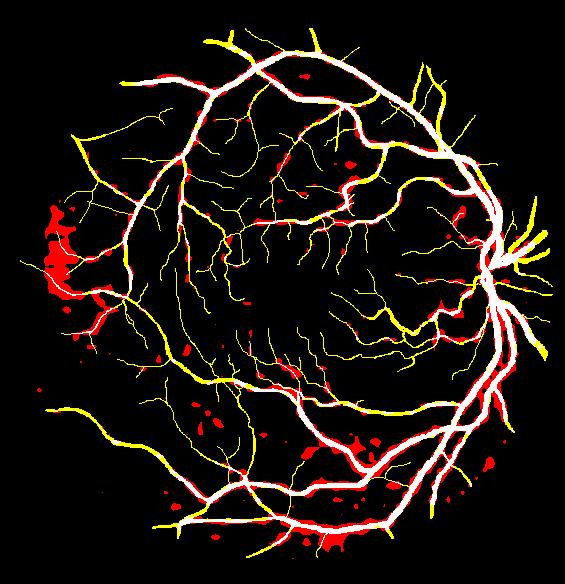}} &
		    \subfigure{\includegraphics[scale=0.072]{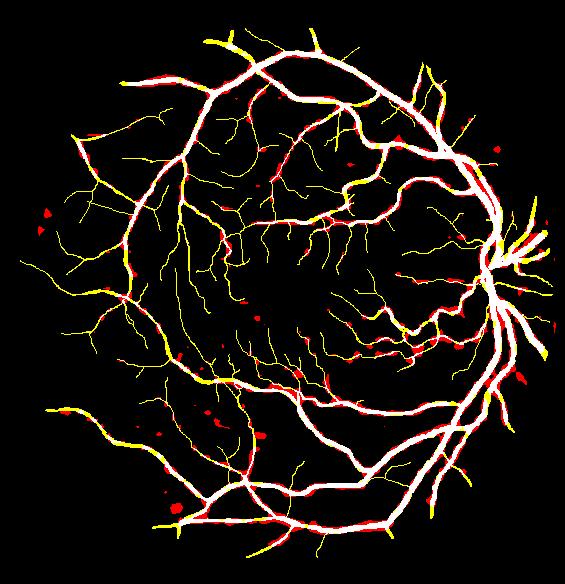}} &
	        \subfigure{\includegraphics[scale=0.072]{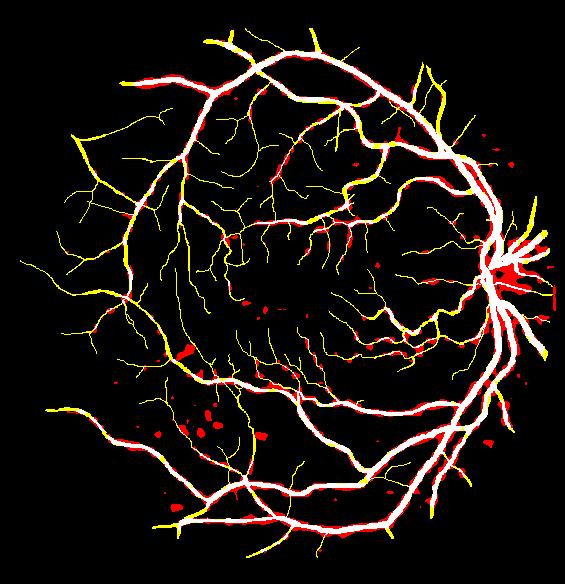}}\\[-1mm]
		\end{tabular}
	\caption{Example visual results for the DRIVE dataset. C1: Input images. C2: Corresponding gold-standard reference segmentations. C3-C5: Output of the model with standard loss and, respectively, training from scratch, pretrained backbone based training, and using additional filters. C6-C8: Output of the model with custom loss and, respectively, training from scratch, pretrained backbone based training, and using additional filters. In C3-C8, white pixels indicate true positives, red pixels indicate false positives, and yellow pixels indicate false negatives.}
	\label{fig:losscomparison}
\end{figure*}

\subsection{Experiment I}
In the first experiment, we used the DRIVE dataset to evaluate the efficacy of the proposed loss function and the corresponding prunability (potential parameter saving) on the lightweight network MobileNetV3-Small \cite{mobileNet}. For a thorough evaluation, variants of the MobileNetV3-Small architecture with unique training settings were compared. These included training from scratch, training from scratch with additional filters added to the backbone, and pretrained backbone. The lightweight network was trained with both the binary cross-entropy loss (default) and the proposed loss (termed as custom loss) to obtain six variants. From the results (Table~\ref{tab:losscomparison}) it is evident that the proposed custom loss achieved performance improvements over the cross-entropy loss in most cases. This confirms the sparsification property of the proposed loss and its ability to suppress unwanted information and attend to important regions of interest. The proposed loss incited parameter savings and train/test time savings (Table~\ref{tab:parameterSaving}). Visual comparison of sample results with the different losses (Fig.~\ref{fig:losscomparison}) illustrates overall superior performance of the proposed loss, in that it resulted in less false pixel predictions.

\subsection{Experiment II}
In the second experiment, we evaluated the performance of the proposed pruning scheme using U-Net~\cite{U-Net} and MobileNetV3-Small~\cite{mobileNet} on the CamVid and DUT-OMRON datasets in terms of mean IOU (mIOU) and the number of model parameters.

\begin{figure}[!t]
\centering
	\includegraphics[scale=0.3]{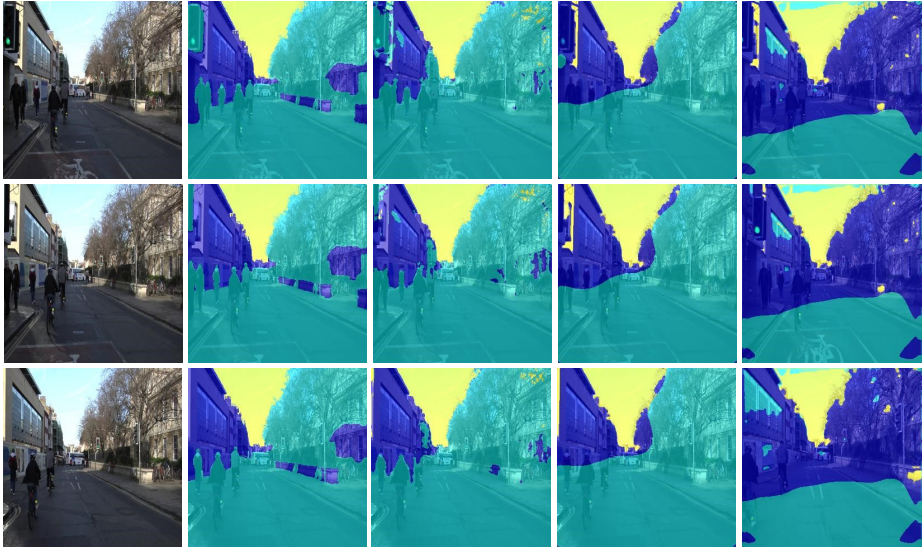}
\caption{Example visual results of the proposed scheme on the CamVid dataset. Columns from left to right: input images, gold standard reference images, U-Net results, U-Net (iter 3) 10\% results, and U-Net (iter 3) 50\% results.}
\label{fig:camvid}
\end{figure}

The results on the CamVid dataset (Table~\ref{tab:camvid}) show a decrease in mIOU for both 10\% and 50\% weight pruning. The drop in performance is relative to the degree of pruning, which could be attributed to training-based damage and network forgetting. For 10\% pruning, there is about 14\% decrease in mIOU after three iterations, with a decrease of about 18\% (4.6 M) parameters. For 50\% pruning, while there is a considerable decrease of about 25\% in mIOU, this is achieved at a significant decrease of about 88\% (22.7 M) parameters. Example visual results (Fig.~\ref{fig:camvid}) show that the segmentation outputs of the 10\% pruned U-Net (iter 3) have considerable overlap with the gold standard and closely follow the original U-Net for the ``building'' class with slight misclassification for the ``sky'' class. These segmentations are achieved with around 18\% fewer parameters compared with U-Net. The U-Net (iter 3) 50\% model still remembers the semantic level segregation between the ``building'' and ``sky'' class but slightly forgets the representation for the background class. It is noteworthy that the 50\% pruned model predicts the building occluded by the trees which is also neglected in the reference maps.

\begin{table*}[!b]
	\centering
	\caption{Pruning results on the CamVid dataset.}
	\begin{tabular}{@{}c@{\hspace{1em}}c@{\hspace{1em}}c@{\hspace{2em}}c@{\hspace{1em}}c@{}}
    \toprule
	\textbf{Dataset} & \textbf{Weights (\%)} & \textbf{Methods} & \textbf{mIOU} & \textbf{\#Params (M)}\\
    \midrule
	\multirow{8}[3]{*}{\begin{sideways}CamVid\end{sideways}} &
	\multirow{4}[2]{*}{10\%} & U-Net & 0.8798 & 25.8 \\
		& & U-Net (iter 1) & 0.8714 & 24.3 \\
		& & U-Net (iter 2) & 0.7822 & 22.8 \\
		& & U-Net (iter 3) & 0.7575 & 21.2 \bigstrut[b]\\
	\cline{2-5} &
	\multirow{4}[1]{*}{50\%} & U-Net & 0.8798 & 25.8 \bigstrut[t]\\
		& & U-Net (iter 1) & 0.6838 & 18.2 \\
		& & U-Net (iter 2) & 0.6087 & 10.6 \\
		& & U-Net (iter 3) & 0.6553 & 3.1 \\
    \bottomrule
	\end{tabular}
	\label{tab:camvid}
\end{table*}

\begin{figure}[!t]
\centering
	\includegraphics[scale=0.33]{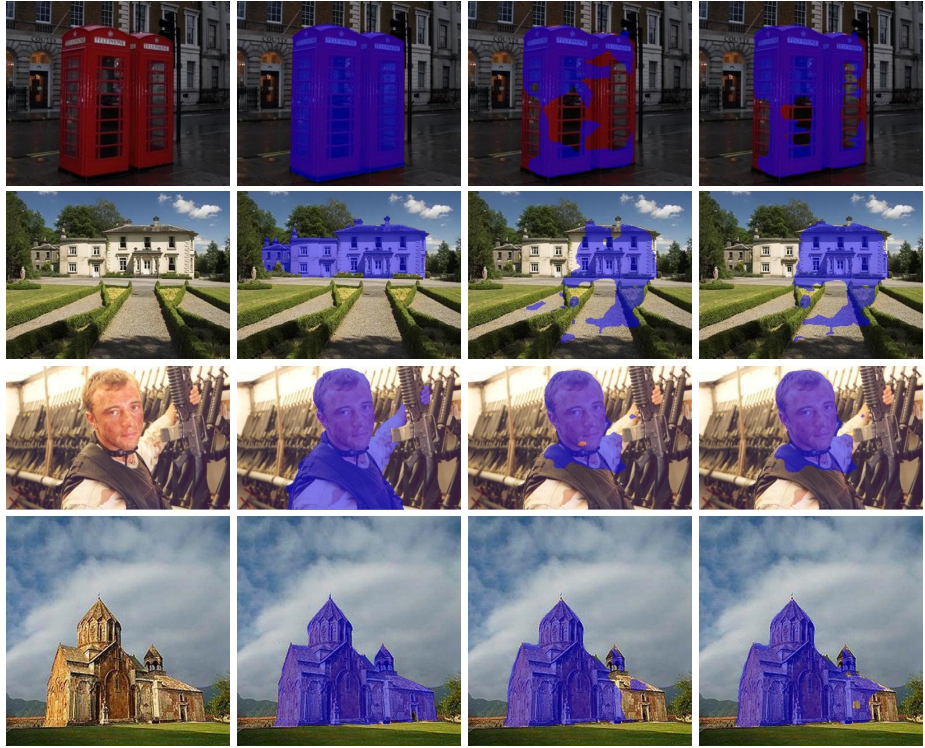}
\caption{Example visual results of the proposed scheme on the DUT-OMRON data\-set. Columns from left to right: input images, gold standard reference images, MobileNetV3-Small results, and MobileNetV3-Small (iter 1) 10\% results.}
\label{fig:dut}
\end{figure}

Results of the proposed pruning scheme for the DUT-OMRON dataset (Table~\ref{tab:dut}) exhibit a slight decrease of about 4\% in mIOU performance of U-Net for 10\% pruning with a 3 M parameter saving. For the 50\% pruned U-Net model we observe about 16\% loss in mIOU at almost 59\% parameter saving. It is noteworthy that the performance of the 10\% pruned MobileNetV3-Small network is improved with a decrease in the number of parameters. Similar to U-Net, there is a considerable decrease in the mIOU for the 50\% pruned MobileNetV3-Small model, but with only a slight decrease of 4.4\% parameters in this lightweight network. Representative visual example results of MobileNetV3-Small on the DUT-OMRON dataset (Fig.~\ref{fig:dut}) show that the pruned model can obtain better coverage of objects of interest. This can be attributed to the prunability of the model and the weight selection of the proposed scheme, which enable the network to better attend to regions of interest.

\begin{table}[!b]
	\centering
	\caption{Pruning results on the DUT-OMRON dataset.}
	\begin{tabular}{@{}c@{\hspace{1em}}c@{\hspace{1em}}c@{\hspace{2em}}c@{\hspace{1em}}c@{}}
    \toprule
	\textbf{Dataset} & \textbf{Weights (\%)} & \textbf{Methods} & \textbf{mIOU} & \textbf{\#Params (M)}\\
    \midrule
    \multirow{10}[12]{*}{\begin{sideways}DUT-OMRON\end{sideways}} & & U-Net & 0.8145 & 25.8 \\
	\cline{2-5} &
	\multirow{2}[2]{*}{10\%} & U-Net (iter 1) & 0.8036 & 24.3 \bigstrut[t]\\
		& & U-Net (iter 2) & 0.7804 & 22.8 \bigstrut[b]\\
	\cline{2-5} &
	\multirow{2}[2]{*}{50\%} & U-Net (iter 1) & 0.6773 & 18.2 \bigstrut[t]\\
		& & U-Net (iter 2) & 0.6843 & 10.6 \bigstrut[b]\\
	\cline{2-5} &
		& MobileNet-V3-Small & 0.4951 & 0.45 \bigstrut\\
	\cline{2-5} &
	\multirow{2}[2]{*}{10\%} & MobileNet-V3-Small (iter 1) & 0.5113 & 0.44 \bigstrut[t]\\
		& & MobileNet-V3-Small (iter 2) & 0.5154 & 0.44 \bigstrut[b]\\
	\cline{2-5} &
	\multirow{2}[2]{*}{50\%} & MobileNet-V3-Small (iter 1) & 0.3005 & 0.43 \bigstrut[t]\\
		& & MobileNet-V3-Small (iter 2) & 0.3231 & 0.43 \\
	\bottomrule
	\end{tabular}
	\label{tab:dut}
\end{table}

\section{Conclusion}
In this paper, a novel training scheme based on composite constraints is presented. The scheme removes redundant filters and reduces the impact that these filters have on the overall learning of the network by promoting sparsity. In addition, in contrast to previous works that make use of pseudo-norm-based sparsity-inducing constraints, the framework that we have developed includes a sparse scheme that is based on gradient counting. The proposed strategy is evaluated on three publicly available dataset across diverse application such as DRIVE (medical image segmentation), DUT-OMRON (saliency detection) and CamViD (scene understanding). The efficacy of the proposed filter pruning is tested on UNet and several variants of MobileNetV3.


\end{document}